\def\year{2020}\relax
\documentclass[letterpaper]{article} % DO NOT CHANGE THIS
\usepackage{aaai20}  % DO NOT CHANGE THIS
\usepackage{times}  % DO NOT CHANGE THIS
\usepackage{helvet} % DO NOT CHANGE THIS
\usepackage{courier}  % DO NOT CHANGE THIS
\usepackage[hyphens]{url}  % DO NOT CHANGE THIS
\usepackage{graphicx} % DO NOT CHANGE THIS
\usepackage{graphicx}  % DO NOT CHANGE THIS
\usepackage{algorithm}
\usepackage{algpseudocode}

\usepackage{subcaption}
\usepackage{enumitem}
\usepackage{amsmath}

\usepackage{booktabs}
\usepackage{tabularx}
\usepackage[table]{xcolor}
\definecolor{Gray}{gray}{0.85}

\title{Tracing Topic Transitions with Temporal Graph Clusters}
\author{{Xiaonan Jing, Qingyuan Hu, Yi Zhang, Julia Taylor Rayz} \\ % All authors must be in the same font size and format. Use \Large and \textbf to achieve this result when breaking a line
Department of Computer and Information Technology \\ %If you have multiple authors and multiple affiliations
Purdue University\\ 
West Lafayette, IN, USA \\
\{jing, hu528, zhan3050, jtaylor1\}@purdue.edu % email address must be in roman text type, not monospace or sans serif
}
 
\begin{document}

\maketitle

\begin{abstract}
Twitter serves as a data source for many Natural Language Processing (NLP) tasks. It can be challenging to identify topics on Twitter due to continuous updating data stream. In this paper, we present an unsupervised graph based framework to identify the evolution of sub-topics within two weeks of real-world Twitter data. We first employ a Markov Clustering Algorithm (MCL) with a node removal method to identify optimal graph clusters from temporal Graph-of-Words (GoW). Subsequently, we model the clustering transitions between the temporal graphs to identify the topic evolution. Finally, the transition flows generated from both computational approach and human annotations are compared to ensure the validity of our framework. 
\end{abstract}

\section{I. Introduction}
\label{sec:intro}
Continuously updating data streams make it challenging to identify real-time topics from platforms like Twitter. Previously, topic identification has mainly been studied on static dataset \cite{stoyanov2008topic,lo2017unsupervised,pappagari2018joint}. However, oftentimes, real-world events are dynamic. During a continuously evolving event, the center of topic can shift as new information being updated throughout the event duration. We are interested in learning the underlying structure of how an event unfolds in the online community, especially when the data are limited for studying user behaviors. 

Stream based event detection aims to identify a sequence of temporal states of the event(s). Given a continuous event across a set of timepoints $\{t_1, t_2, ..., t_n\}$, we define a temporal (sub)-event as the state $s_i$ of the event at any timepoint $t_i (i < n)$, with $t_n$ being the final timepoint whereas the event has no further updates. One of the biggest challenges in stream based detection lays in locating the temporal state boundary across the timepoints. When a dynamic event evolves over time, the temporal state $s_1$ may remain unchanged across several timepoints, however, suddenly converting to state $s_2$ then subsequently emerging to $s_3$. In other words, the temporal state at each timepoint is not independent of each other. There is a transition between the states when a change is to occur. Thus, the detection of a significant change in state becomes crucial in stream based tasks. Previously, burst based detection \cite{mcminn2015real,kaneko2016event} and anomaly based detection \cite{guille2015event,fedoryszak2019real} have been explored. Burst based detection exploits a frequency based approach, which does not emphasize the semantic content of the events. On the other hand, anomaly based detection focuses on the change of semantic topic in textual content. In this paper, we are interested in the latter type, which traces the semantic topic change in a continuous time space. 

We propose to employ graph structure to model temporal states due to its flexibility in relationship assignment and scalability in computational cost. Graphs have been adopted for similar tasks in event identification \cite{meladianos2015degeneracy,jing2020graph}. A graph $G= (V, E)$ generally consists of a set of vertices $V$ and a set of edges $E$ which connects the vertices. The vertices can be words, sentences, or documents, and the edges can be used to model the statistical or semantic relationships between the textual elements. Our approach utilizes Graph-of-Words (GoW) to construct a temporal graph for tweets content at each timepoint, allowing the graph clustering to group temporal topics. Consequently, we develop topic transition flow by modeling cluster transitions at global level across all timepoints. Our main contributions are: 1) development of a clustering with nodes removal algorithm to find the optimal graph clusters over topical dataset; 2) improvement on cluster transition modeling to simulate the transition across all timepoints as well as taking re-emerging clusters into consideration; 3) visualization of topic transition flows in the time space. 

    \begin{table*}[t]
        \caption{Statistics of the dataset split by day}
        \centering
        \small %%% change text to small size
        \begin{tabular}{lccccccccccccccc}
            \hline
            Timepoint & \multicolumn{1}{l}{8/19} & \multicolumn{1}{l}{8/20} & \multicolumn{1}{l}{8/21} & \multicolumn{1}{l}{8/22} & \multicolumn{1}{l}{8/23} & \multicolumn{1}{l}{8/24} & \multicolumn{1}{l}{8/25} & \multicolumn{1}{l}{8/26} & \multicolumn{1}{l}{8/27} & \multicolumn{1}{l}{8/28} & \multicolumn{1}{l}{8/29} & \multicolumn{1}{l}{8/30} & \multicolumn{1}{l}{8/31} & \multicolumn{1}{l}{9/1} & \multicolumn{1}{l}{9/2} \\ \hline
            Tweets & 38 & 89 & 87 & 65 & 27 & 68 & 53 & 29 & 19 & 53 & 23 & 18 & 40 & 35 & 16 \\
            Nodes & 139 & 218 & 167 & 177 & 69 & 189 & 204 & 101 & 65 & 137 & 72 & 56 & 122 & 128 & 113 \\ \hline
            \end{tabular}
        \label{tab:dataset}
    \end{table*}

\section{II. Related Work} % by yi, not proof read yet
\label{sec:lit}
In this section, we review previous work on event identification on Twitter and graph based event modeling. 

% event clustering
% Jin and Bai 2016
We start with event clustering graphs, for which Jin and Bai (\citeyear{jin2016clustering}) proposed a long document clustering approach utilizing a directed GoW for representing the word features contained in each document. 
%Each document was translated into a graph describing word features, co-occurrence, and co-occurrence frequency, respectively, with nodes, edges, and edge weights. 
The document clusters were generated based on the maximum common subgraphs between each document graph. 
% edouard et al 2017
In a similar attempt, Edouard et al. (\citeyear{edouard2017graph}) proposed an event clustering model which leveraged named entities (NE) based directed GoW structure. The GoW was improved by using surrounding context of the graph nodes, NE, to enrich node level information. The approach is capable to automatically detect different events with the same keywords without any prior knowledge. 
% jinarat et al 2018
Jinarat et al. (\citeyear{jinarat2018clustering}) employed a GoW combined with pretrained Word2Vec embedding \cite{word2vec} for tweet clustering. The Word2Vec similarity served as a metric for edge removal in generating tweet clusters. However, since abbreviations and hashtags occur regularly in tweets, pre-trained embeddings can be vulnerable to these irregularities which may not present in the training data. 

% stream based event detection 
Extracting event streams of an ongoing event from Twitter has a goal of detecting how an event unfolds as people post updates.
% meladianos et al 2015
Meladianos et al. (\citeyear{meladianos2015degeneracy}) improved the GoW approach by integrating the tweet length with the global co-occurrence frequency to identify the sub-events of a World Cup match on Twitter. Tweets containing the top k degenerated subgraph were used to describe each sub-event.
% fedoryszak et al 2019
Fedoryszak et al. (\citeyear{fedoryszak2019real}) proposed an interpretation of Twitter event streams - a chain of clustered trending entities arranged in chronological order. Additionally, Fedoryszak et al. overcame the limitation of lack of coverage in event detection with the aid of Twitter's internal knowledge graph (KG).
% jing and rayz 2020
Jing and Rayz (\citeyear{jing2020graph}) introduced Graph of Tweets (GoT) for modeling popular events with both word and document level structures. GoT treat a tweet as a collection of conceptualized token nodes, whereas tokens of contextual similarities were merged prior to the GoT construction. Popular sub-events were extracted by detecting cliques among the similar tweet nodes.% characterized by the normalized mutual information (MI). 

% event definition on twitter 
Last but not least, we review previously proposed event representation on Twitter as there has not been a formal definition due to the various nature of the tasks. Hashtag based event identifications \cite{feng2015hashtag,yang2018hashtag} treat an event as "a group of hashtags that focus on the same topic". 
Another approach utilizes NE to define event, whereas each NE is treated as an individual event and a set of NE as a merged event \cite{mcminn2015real}. 
Text triplet (subject, predicate, object) has also been adopted to describe a Twitter event \cite{tonon2017triplet}. In such approaches, OpenIE \cite{angeli2015} has been one of the top candidates for triplet extraction. 
Finally, embedding based approach which treats each tweet embedding as an event has also been explored \cite{dhingra2016tweet2vec}. 
All of the above representations have their pros and cons -- hashtags are excellent carries of topical information, but they may not be present in every tweet; NE can support details of the event, but they may also cause crucial information to be excluded (i.e. pronouns which are often subjects of an event); triplets can provide relational information, but entities that are not involved in a triplet cannot be captured; Embeddings allow efficient mathematical computations but also requires an adequate amount of data to train. 
We combine hashtags, NE and nouns in this paper to represent Twitter topics due to the limited amount of data we could collect for our experiment. 

\section{III. Proposed Method}
\label{sec:method}
We are interested in identifying topic transitions in specified events. We break the task into the following steps:
 1) constructing a temporal graph for each timepoint; 2) applying clustering with node removal on each temporal graph; 3) modeling cluster transition flow across timepoints. 

% data, and dataset split
\subsection{Dataset}
\label{sec:dataset}
Opportunistically, we chose to model local responses to the on-going event "COVID-19" for a short duration. Thus, tweets were collected from a local community from Aug 19th to Sep 2nd.  "COVID-19" related tweets were identified by matching a set of manually selected hashtags for the corpus. The tweets of interests are pre-processed with Stanford CoreNLP \footnote{Stanford CoreNLP ver. 3.9.2.} to annotate the part-of-speech and named entities. The statistics of the dataset is shown in Table \ref{tab:dataset}.

% graph construction
\subsection{Graph Construction} 
\label{subsec:graph_construction}
We treat each day as a timepoint and split the collected dataset based on the timestamp of the tweets. Temporal graphs are constructed using GoW with the nodes being the unique nouns or named entities extracted from tweets of the day. Furthermore, we employ normalized Point-wise Mutual Information (PMI) value as the edge weights between two nodes (\ref{eq:pmi}). In Equation (\ref{eq:pmi}), the marginal probabilities $p(x)$ and $p(y)$ and the joint probability $p(x, y)$ are computed as the proportions of the occurrence of tokens $x$ and $y$ in a total of $N$ tweets, where $n_x$, $n_y$, and $n_{xy}$ denote the (joint) frequency of tokens $x$ and $y$. For consistency and computational efficiency, we further normalize the PMI with self-information $h(x, y)$ which sets the boundary of the PMI value to $[-1, 1]$ (\ref{eq:npmi}). 
% pmi functions
    \begin{align}
        % pmi 
        \begin{split}
            \label{eq:pmi}
            pmi &= log\frac{p(x, y)}{p(x)p(y)} = log\frac{n_{xy}}{n_x n_y} N
        \end{split}\\
        % npmi 
        \begin{split}
            \label{eq:npmi}
            npmi &= \frac{pmi}{h(x, y)} = \frac{pmi}{-log p(x, y)} 
      \end{split}
    \end{align}

% markov clustering of graph
\subsection{Clustering with Node Removal} 
\label{subsec:clustering} 
%GoW constructed for a topical dataset will likely contain several bridging nodes. 
For the dataset used in this paper, nodes that are closely related to the fetching keywords of the tweets tend to co-occur with every other node, while their neighboring nodes might observe no connections between each other. We refer to this type of nodes as the bridging nodes in this paper. Many common graph clustering methods, such as spectral clustering \cite{ng2002spectral} and highly connected subgraph clustering \cite{hartuv2000clustering}, achieve the grouping from graph cutting. One drawback of applying cutting based algorithm  on graphs with bridging nodes is that no obvious local structures can be observed due to the inter-connectivity introduced by these nodes. More precisely, the cluster assignments for the neighboring nodes of a bridging node tend to fail as the graph cannot be cut in an appropriate way. As a result, the graph cutting mechanism tends to cluster each node into individual cluster. 
Thus, we propose to exclude the bridging nodes from the GoW during the clustering process, and treat them as belonging to each resulting clusters which have at least one edge in between. We locate a bridging node by its clustering coefficient $C_i \in [0, 1]$ \cite{watts1998collective}. As defined in Equation \ref{eq:clustering_coefficient}, clustering coefficient measures the embeddedness of a single nodes among its neighbors. A larger $C_i$ indicates the neighbors of $i$ tend to be more connected to form a community. In our case, the smaller the $C_i$ is, the more likely the node is to serves as bridging node. 

    % clustering coefficient function
    \begin{equation}
    \label{eq:clustering_coefficient}
        C_i = \frac{2e_i}{k_i(k_i - 1)}
    \end{equation}
    % modularity function 
    \begin{equation}
    \label{eq:modularity}
        Q = \frac{1}{2m} \sum_{ij} (A_{ij} - \frac{k_ik_j}{2m}) \delta(c_i, c_j)
    \end{equation}
    % function notations 
    \begin{itemize}[nosep]
        \item $e_i$: number of edges between the neighbors of node $i$
        \item $k_i$ and $k_j$: degree of node $i$ and $j$
        \item $A_{ij}$: the edge weight between nodes $i$ and $j$
        \item $2m$: sum of all edge weights 
        \item $c_i$ and $c_j$: communities of nodes $i$ and $j$
        \item $\delta$: an indicator function, $\delta = 1$ if $c_i = c_j$, $\delta = 0$ otherwise
    \end{itemize}
    
To achieve an optimal clustering, we determine the number of bridging nodes to exclude by incrementally removing the denser node from the graph based on a clustering quality metric. Modularity \cite{newman2006modularity} is a common metric used for measuring community quality in graph theory. Given a partitioning of graph $G$, modularity $Q \in [-1, 1]$ computes the difference between actual and expected number of edges within groups (Equation \ref{eq:modularity}). A larger modularity value $Q$ indicates more significant community structure. Algorithm \ref{alg:optim_cl} outlines our method for finding optimal clustering. During each iteration, a node with the lowest $C_i$ is removed (with its edges) from the graph $G$ and the rest of the subgraph is clustered. The modularity value $Q$ is computed on the subgraph to determine the current clustering quality. The best clustering is achieved at the (first) max $Q$ value. 

    % clustering algorithms with nodes removal
    \begin{algorithm} [ht]
    \begin{algorithmic}
    \caption{Finding Optimal Graph Clustering}
    \label{alg:optim_cl}
        % \Function{FindOptimalGraphClustering}{Graph $G$}
        \For{node $v_i \in V:\{v_1, v_2, ..., v_m\}$} 
            \State $C_i = \textit{\textbf{Equation\_3}}(v_i)$
        \EndFor
        \State $ V = \textit{\textbf{sort\_asc}}(V, key=C_i) $ 
        \State $ best\_clustering = \textbf{None} $
        \State $ Q\_max = -1 $
        \For{$v_i \in V$} 
            \State $ G.\textit{\textbf{remove}}(v_i) $
            \State $ clusters = \textit{\textbf{Clustering}}(G) $
            \State $ Q = \textit{\textbf{Equation\_4}}(clusters) $
            \If{ $ Q > Q\_max $} 
                \State $ Q\_max = Q$
                \State $ best\_clustering = clusters $
            \EndIf
        \EndFor 
        \State \Return $best\_clustering$
        % \EndFunction
    \end{algorithmic}
    \end{algorithm}

% markov clustering choice justification 
We adopt the random walk based Markov Clustering (MCL) \cite{van2008graph} as our choice of graph clustering algorithms over other common graph cutting based clustering algorithms due to the drawbacks mentioned above. MCL simulates the stochastic flow in a graph which makes it more scalable. Furthermore, unlike other clustering algorithms, the number of clusters does not need to be pre-determined in MCL. 

    % cluster transition types 
    \begin{table*}[t]
        \centering
        \caption{Cluster transition types for cluster $X$ at timepoint $t_i$}
        \begin{tabular}{|c|c|} \hline \\[-1em]
            \textbf{Transition Type} & \textbf{Mathematical Definition} \\ \hline \\[-1em]
            the cluster stays unchanged & $X \rightarrow X'$, where $X' = match_{\alpha}(X)$ \\ \hline \\[-1em]
            the cluster is absorbed & $X \rightarrow Y$, where $match_{\alpha}(X) \subset Y$ and $X-match_{\alpha}(X) \not\subset Y $\\ \hline\\[-1em]
            the cluster is dissolved & $X \rightarrow Y$, where $match_{\alpha}(Y) \subset X$ and $Y-match_{\alpha}(Y) \not\subset X$\\ \hline \\[-1em]
            the cluster splits into multiple clusters & $X \rightarrow \{{Y_1, Y_2, ..., Y_m}\}$, where $\bigcup_{1}^{m} Y_j = match_{\alpha}(X)$ \\ \hline \\[-1em]
            the cluster is merged from multiple clusters & $\{{X_1, X_2, ..., X_n}\} \rightarrow Y$, where $\bigcup_1^n X_i = match_{\alpha}(Y)$\\ \hline\\[-1em]
            the cluster disappears & $X \rightarrow$ \O \\ \hline\\[-1em]
            a new cluster has emerged & \O\ $ \rightarrow Y$ \\ \hline\\[-1em]
            a cluster has re-emerged & $X \rightarrow$ \O $\rightarrow X'$, where $X' = match_\alpha (X)$ \\ \hline
        \end{tabular}
        \label{tab:cluster_transition_types}
    \end{table*}
    
    % cluster metadata
    \begin{table*}[t]
        \caption{Statistics of the GoW During Clustering}
        \centering
        \small %%% change text to small size
        \begin{tabular}{@{}lccccccccccccccc@{}}
        \toprule
        Timepoint & \multicolumn{1}{l}{8/19} & \multicolumn{1}{l}{8/20} & \multicolumn{1}{l}{8/21} & \multicolumn{1}{l}{8/22} & \multicolumn{1}{l}{8/23} & \multicolumn{1}{l}{8/24} & \multicolumn{1}{l}{8/25} & \multicolumn{1}{l}{8/26} & \multicolumn{1}{l}{8/27} & \multicolumn{1}{l}{8/28} & \multicolumn{1}{l}{8/29} & \multicolumn{1}{l}{8/30} & \multicolumn{1}{l}{8/31} & \multicolumn{1}{l}{9/1} & \multicolumn{1}{l}{9/2} \\ \midrule
        $\bar{C}_{rm}$ & 0.41 & 0.47 & 0.47 & 0.48 & 0.46 & 0.48 & 0.39 & 0.74 & 0.33 & 0.41 & 0.43 & 0.4 & 0.44 & 0.42 & 0.28 \\
        $\bar{C}_{all}$ & 0.95 & 0.89 & 0.93 & 0.91 & 0.93 & 0.92 & 0.93 & 0.96 & 0.95 & 0.92 & 0.94 & 0.93 & 0.94 & 0.96 & 0.96 \\
        $\bar{C}_{best}$ & 0.99 & 0.99 & 0.99 & 0.99 & 0.99 & 0.99 & 0.99 & 1.0 & 0.99 & 0.99 & 0.98 & 0.98 & 0.98 & 0.996 & 0.98 \\
        $\%_{rm}$ & 7.19 & 18.81 & 11.98 & 16.95 & 11.59 & 14.81 & 10.78 & 16.83 & 6.15 & 12.41 & 6.94 & 8.93 & 8.2 & 5.47 & 3.54 \\ \bottomrule
        \end{tabular}
        \label{tab:clustering_metadata}
    \end{table*}

% clustering transition 
\subsection{Modeling Cluster Transitions} 
\label{subsec:cluster_transition} 
In many stream based data analysis tasks, modeling cluster transitions is the key to identify the evolution of the target of interests. Given a timepoint $t_i$, a cluster transition can be defined as "the change experienced by a cluster that has been discovered at an earlier timepoint" \cite{spiliopoulou2006monic}. Previously proposed frameworks such as MONIC \cite{spiliopoulou2006monic} and MEC \cite{oliveira2012framework} define a set of transition types between clusters across consecutive timepoints and use a matching function with a threshold to identify these types. We adopt the basic transition types defined in MONIC and MEC, and further extend them to make the framework more robust for our task. From timepoint $t_i$ to $t_{i+1}$, we define pairwise transition types for consecutive timepoints in Table \ref{tab:cluster_transition_types} (first 7 types), where $X$ and $Y$ are clusters at timepoints $t_i$ and $t_{i+1}$ respectively, $\alpha$ is the threshold for the $match$ function to measure the overlaps between two clusters. In addition to transition types defined between pairs of consecutive timepoints, we introduce another transition type, namely, "a cluster has re-emerged", which measures the transition across non-consecutive timepoints. To visualize the transitions of a cluster in a global view, we model the set of transitions starting at a newly emerged cluster as a flow. In other words, we treat the pairwise transitions as the edges between the cluster nodes across different timepoints. When a transition exists between two cluster nodes, an edge with the transition type as value is assigned to connect them. It should be noted that to be computationally consistent with the basic transition types, the re-emergence transition is matched in respect of the last node in the sequence of the consecutive transitions.

\subsection{Evaluation}
\label{subsec:eval}
We validate the topic transition results from our computer generated approach by applying the same transition framework to human annotated clusters. Each tweet is annotated with at most three noun (-phrase) labels which summarize the tweet the most by a human expert. The majority of the labels are directly selected from the tweets. Out-of-content labels are only generated when the tokens in a tweet cannot meaningfully summarize it.

A preliminary inspection over the manually and computationally generated clusters showed a mismatch in quantity of labels. In addition, we observed several co-occurring topics in the annotated clusters due to retweets. To merge the co-occurring topics in human annotated clusters, we employ frequent itemsets \cite{hornik2005arules} to discover strongly associated topics. Support $supp$ is calculated as an indication of how frequently an itemset appears in the data (Equation \ref{eq:support}). If two itemsets share the same $supp$ value and one itemset is the subset of another, we merge the subset into its parent.
    
    \begin{equation}
    \label{eq:support}
        supp(X) = \frac{|t \in T; X \subseteq t|}{|T|}
    \end{equation}

    \begin{itemize}[nosep]
        \item $T$: a set of transactions of a given dataset.
    \end{itemize}

It should be noted that the methods of clustering between computer generated and human-annotated data are very different. However, the trends should be visible regardless of the methods, provided that the clusters are done well. 

\section{IV. Experimental Results}
\label{sec:result}
The number of nodes distributed in each temporal GoW separated by timepoint is shown in Table \ref{tab:dataset}. As previously mentioned, we are interested in learning the responses and interests of a local community, and the real-world data we collected is limited in both quantity and quality. Thus, the number of nodes in each GoW varies largely depending on the community's activity of that day. Reporting global average based results would not guarantee an accurate evaluation to our framework. We instead will report results by timepoint and evaluate our framework by trend based comparison between computer and human generated clusters to ensure the validity of this work. 

    % figure: example cluster result 0819
    \begin{figure}[hbt!]
        \centering
        \includegraphics[width=\linewidth]{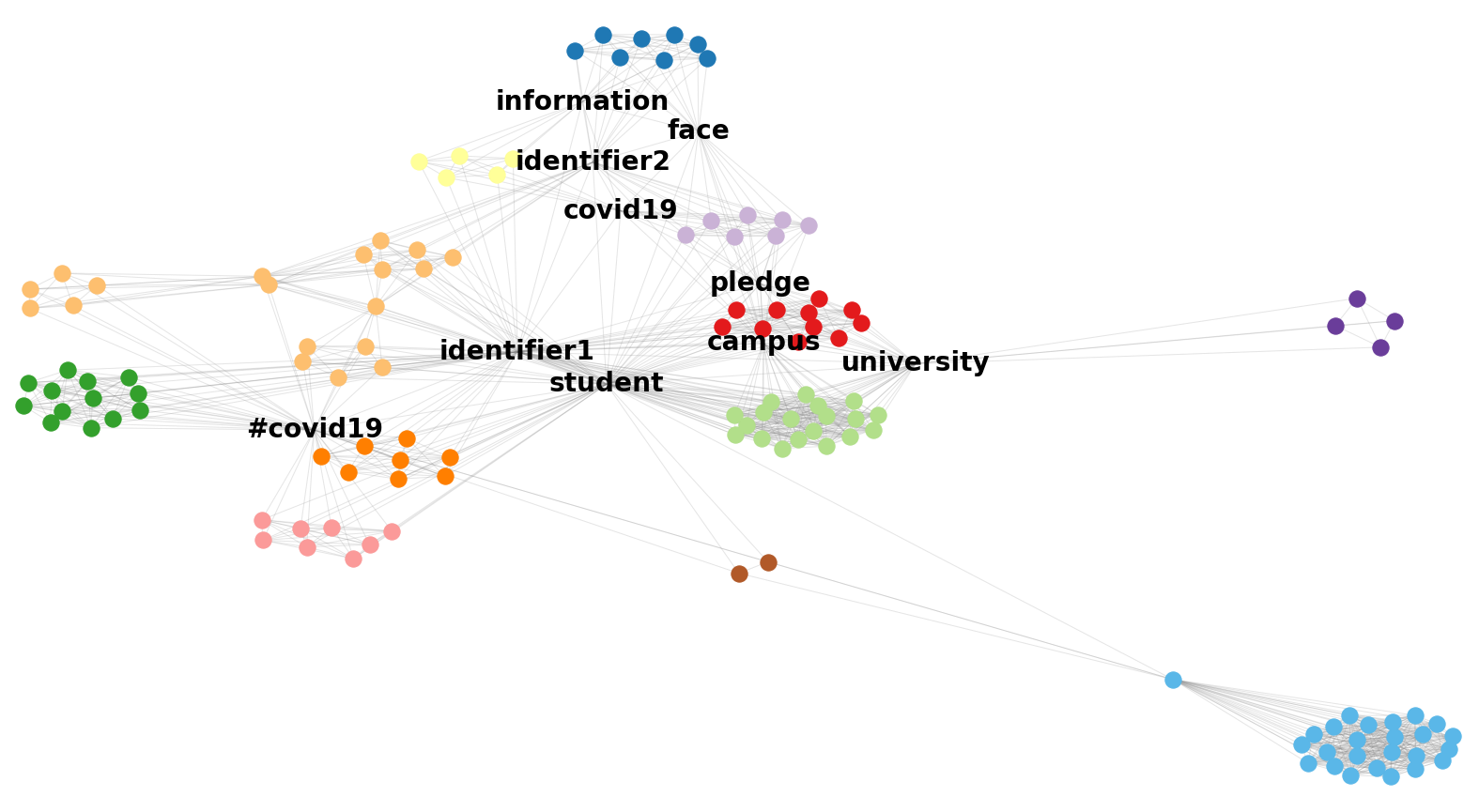}
        \caption{An instance of event graph for Aug 19, 2020 with 12 clusters. Each color represents a cluster and the bold (anonymized) words are the bridging nodes.}
        \label{fig:plot_clusters}
    \end{figure}
    
      % figure: transition visualization 
    \begin{figure*}[hbt!]
    \begin{subfigure}[b]{0.49\textwidth}
        \includegraphics[width=\linewidth]{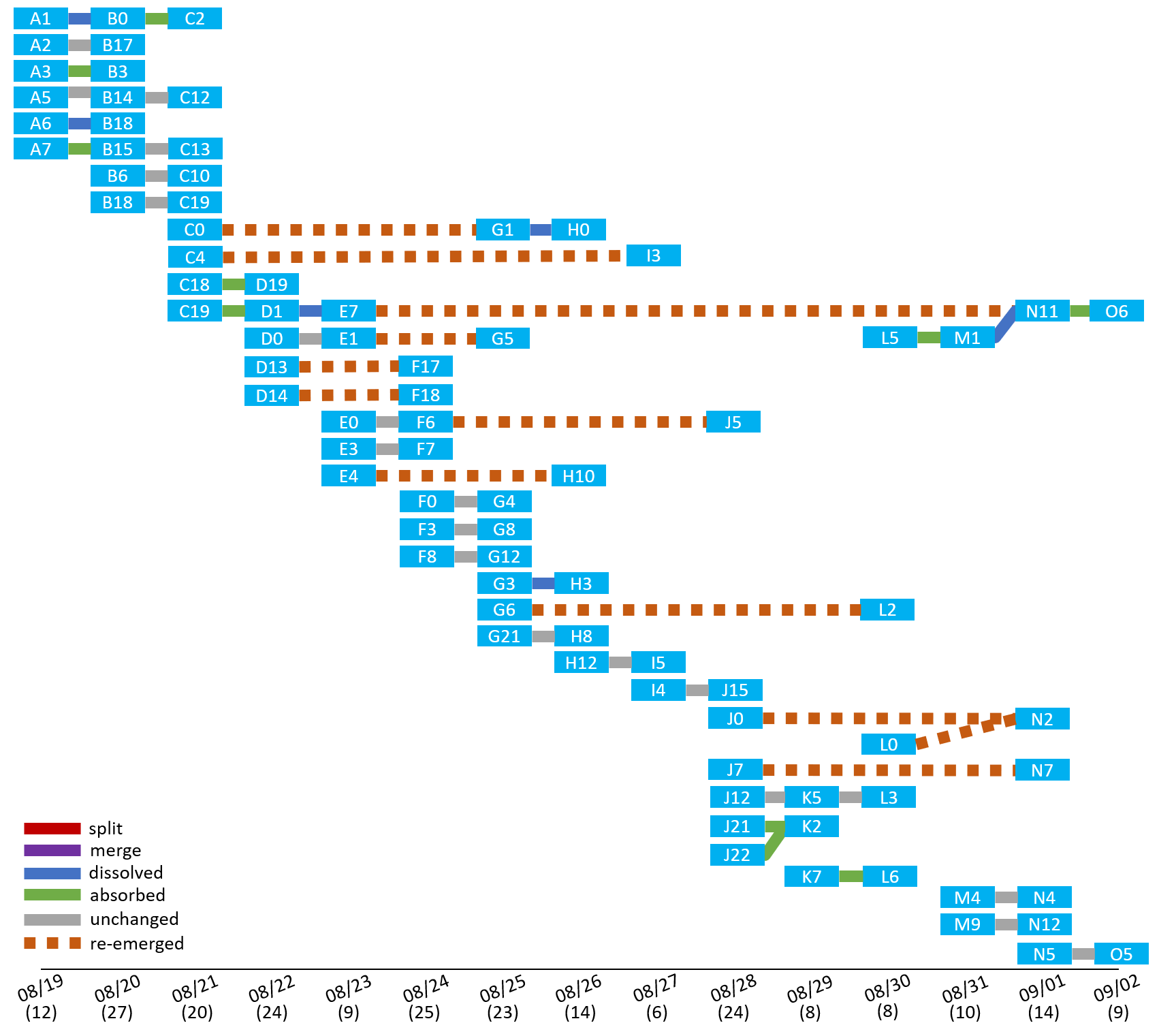}
        \caption{Computed Clusters} \label{fig:transition_a}
    \end{subfigure}%
    \hspace*{\fill}
    \begin{subfigure}[b]{0.49\textwidth}
        \includegraphics[width=\linewidth]{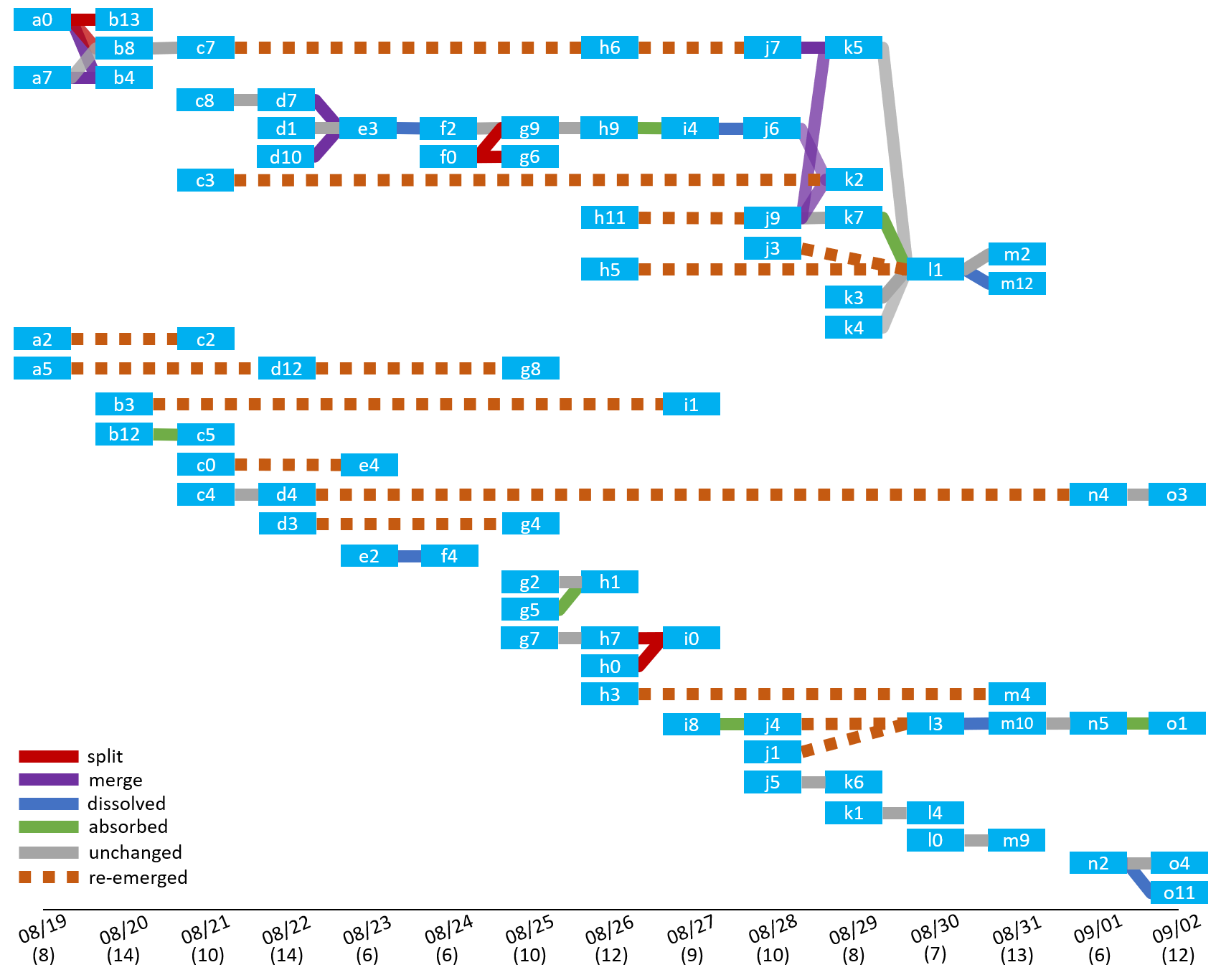}
        \caption{Human Annotated Topics} \label{fig:transition_b}
    \end{subfigure}
    \caption{Cluster progression charts for (a) computed clusters vs. (b) human annotated topics from Aug. 19, 2020 to Sep. 02, 2020. Numbers below each date indicate the total number of clusters for the given date.}
    \label{fig:transition}
    \end{figure*}
\subsection{Clustering with Node Removal}
To summarize, the average percentage of bridging nodes removed in each temporal GoW is 10.71\% with an average $C_i$ of $0.44$. Timepoint specific data of the clusterings is shown in Table \ref{tab:clustering_metadata}, where $\bar{C}_{rm}$, $\bar{C}_{all}$, $\bar{C}_{best}$ denote the average $C_i$ for removed nodes, the original GoW, and the best clustered subgraph respectively; and $\%_{rm}$ denotes the percentage of nodes removed. The best subgraphs across all timepoints showed an increase in average $C_i$ after the removal of bridging nodes, which confirms that the global embeddedness of the graph have become stronger. It is noteworthy that many of the best subgraphs achieve a nearly 1.0 (maximum boundry) average $C_i$, which suggests that the neighbors of each node in the subgraph are inter-connected. This can occur when the resulting subgraph is a complete graph or consists several fully connected components which are not inter-connected. Our results are latter. Another metric should be considered for future tasks as the clustering coefficient $C_i$ lacks the ability to differentiate the structural characteristics since it only considers the internal variance of the clusters. 

In addition, we observed consistent converging trends among the modularity plots over all timepoints. %Figure \ref{fig:plot_modularity} shows an example of the change in modularity values on Aug 19 throughout the process of node removal as defined in Algorithm \ref{alg:optim_cl}. 
As the nodes with lower $C_i$ get excluded from the subgraph, the resulting clustering quality starts to increase and gradually converges. Once a maximal modularity value is achieved, removing more bridging nodes will cause a decay in quality. Currently, our approach chose the result at the maximal modularity value as the optimal clustering. To increase generalizability and reduce computational cost, Algorithm \ref{alg:optim_cl} can be updated to stop when the modularity value starts to converge. Figure \ref{fig:plot_clusters} illustrates an example clustering on Aug 19. 
    % figure: example modularity values 0819
   % \begin{figure}[t]
   %     \centering
    %    \includegraphics[width=\linewidth]{./figures/0819_modularity.png}
    %    \caption{Modularity Plot for Aug 19, 2020. The best clustering is achieved at 10 (red dot) out of 139 nodes removed, resulted in 12 clusters in subgraph.}
   %     \label{fig:plot_modularity}
   % \end{figure}

\subsection{Comparing Cluster Transition}  % unfinished grrr....
We applied the metrics defined in Table \ref{tab:cluster_transition_types} to construct transition flows for both the computer generated clusters and human annotated label sets. An opportunistic threshold of $\alpha = 2/3$ is used for the $match$ function to determine if two clusters are considered as the same, no other threshold have been attempted. The computational approach generated 34 independent transition flows with an average length of 2.41 timepoints per flow, while 151 clusters only exist across a single timepoint. On the other hand, the human annotation suggests that there are 17 independent transition flows with an average length of 4.53 timepoints per flow, and 68 sets only exist across a single timepoint. 

It should be noted that the computer generated clusters are based on nouns and NE from tweets with no order in mind, while human-annotated clusters are based on three summarized terms per tweet. Thus, computer generated clusters group all topic per day together irrespective of which tweet they come from, while human annotated ones have an additional layer of summarization: each tweet to keywords. The mismatch in the number of clusters can thus be explained by this additional layer of complexity, and should not necessarily be treated as a negative result. What is more interesting to see is the patterns that emerge from repeating topics, within both layers, throughout two weeks of data.

Figure \ref{fig:transition} demonstrates the cluster progression charts of the computational approach and human annotated topics side by side. It can be seen that many clusters, both computer and human generated, reappear overtime. 
%Do they share similar topics? if so, list them
One clear outlier is a sequence of human-annotated clusters that contains a series of various transitions for over a week. A shared label between these clusters denotes the name of a policy that the local community is enforcing to ensure safety during the pandemic. Coincidentally, the same policy name has been detected as the bridging node in computer generated clusters in nearly all timepoints during the clustering process.
Further analysis suggests that the longest sequence of transitions in both computer and human generated clusters can be traced to the same topic not mentioned here due to anonymity requirement. 
Despite the differences between the cluster generating mechanisms, both transition flows progress similarly in trends of emergence, re-appearance, and disappearance.

\section{V. Conclusion and Future Work} 
\label{sec:conclusion}
In this paper, we proposed a graph based framework \footnote{The framework implementation details can be found at https://github.com/lostkuma/TopicTransition} in modeling topic transitions of a local online community throughout two weeks of Tweets. %The main components of our framework consist of constructing a temporal GoW for each timepoint, clustering with nodes removal to produce optimal topical clusters, and modeling cluster transitions to generate topic flows. 
Our proposed clustering with node removal approach attempted to resolve the drawback of the traditional hard clustering method on topical datasets. The cluster transition modeling provided insides on how topic flows can be traced in a continuous time space. Finally, the flow comparison between computer generated clusters and human annotated label sets demonstrated the applicability of our framework on real-world data. 

Several improvements can be made in the future: 1) using a convergent schema to locate the best modularity value for the best subgraph to increase the model's robustness and reduce computational costs; 2) the assignments of membership for the bridging nodes to each resulting clusters as currently they are considered to belong to all inter-connected cluster with equal weights; 3) incorporating conceptual level information (i.e. knowledge graph) into the temporal graph to refine the events/topics clusters. However, even without these improvements, our framework is capable of tracing the evolution of the topics and the duration of the progressions as evident by comparison with human annotation.

% References and End of Paper
% These lines must be placed at the end of your paper
\bibliography{references}
\bibliographystyle{aaai}

\end{document}